\documentclass{bmvc2k}

\title{FAST3D: Flow-Aware Self-Training for 3D Object Detectors}

\addauthor{Christian Fruhwirth-Reisinger}{christian.reisinger@icg.tugraz.at}{1,2}
\addauthor{Michael Opitz}{micopitz@amazon.de}{1,2}
\addauthor{Horst Possegger}{possegger@icg.tugraz.at}{2}
\addauthor{Horst Bischof}{bischof@icg.tugraz.at}{1,2}

\addinstitution{
	Christian Doppler Laboratory for\\Embedded Machine Learning
}

\addinstitution{
	Institute of Computer Graphics and Vision\\
	Graz University of Technology
}
\runninghead{Fruhwirth-Reisinger, Opitz, Possegger, Bischof}{FAST3D}

\def\eg{\emph{e.g}\bmvaOneDot}

\def\ie{\emph{i.e}\bmvaOneDot}
\usepackage{multirow}

\renewcommand{\vec}[1]{\ensuremath{\mathbf{#1}}}

\begin{document}

\maketitle

\begin{abstract}	
	In the field of autonomous driving, self-training is widely applied to mitigate distribution shifts in LiDAR-based 3D object detectors. This eliminates the need for expensive, high-quality labels whenever the environment changes (\eg, geographic location, sensor setup, weather condition). State-of-the-art self-training approaches, however, mostly ignore the temporal nature of autonomous driving data. %  which contains additional information.
	
	To address this issue, we propose a flow-aware self-training method which enables unsupervised domain adaptation for 3D object detectors on continuous LiDAR point clouds. In order to get reliable pseudo-labels, we leverage scene flow to propagate detections through time. In particular, we introduce a flow-based multi-target tracker, that exploits flow consistency to filter and refine resulting tracks. The emerged precise pseudo-labels then serve as a basis for model re-training. Starting with a pre-trained KITTI model, we conduct experiments on the challenging Waymo Open Dataset to demonstrate the effectiveness of our approach. Without any prior target domain knowledge, our results show a significant improvement over the state-of-the-art.

\end{abstract}

%-------------------------------------------------------------------------
\section{Introduction}
In order to safely navigate through traffic, self-driving vehicles need to robustly detect surrounding objects (\ie~vehicles, pedestrians, cyclists).
State-of-the-art approaches leverage deep neural networks operating on LiDAR point clouds~\cite{shi2020points, shi2018pointrcnn, shi2020pvrcnn}.
However, training such 3D object detection models usually requires a huge amount of manually annotated high-quality data~\cite{sun2020waymo, caesar2020nuscenes, geiger2012ad}.
Unfortunately, the labelling effort for 3D LiDAR point clouds is very time-consuming and consequently also expensive -- a major drawback for real-world applications.
Most datasets are recorded in specific geographic locations (\eg Germany~\cite{geiger2012ad, geiger2013vision}), with a fixed sensor configuration and under good weather conditions.
Applied to new data collected in other locations (\eg USA~\cite{sun2020waymo}), with a different sensor (\eg sparser resolution~\cite{caesar2020nuscenes}) or under adverse weather conditions (\ie fog, rain, snow), 3D detectors suffer from distribution shifts (domain gap). This mostly causes serious performance drops~\cite{wang2020train}, which in turn leads to unreliable recognition systems. 

This can be mitigated by either manual or semi-supervised annotation~\cite{walsh2020temporallabeling, meng2020ws3d} of representative data, each time the sensor setup or area of operation changes.
However, this is infeasible for most real-world scenarios given the expensive labelling effort.
A more general solution to avoid the annotation overhead is unsupervised domain adaptation (UDA), which adapts a model pre-trained on a label-rich source domain to a label-scarce target domain.
Hence, no or just a small number of labelled frames from the target domain are required.

For 3D object detection, UDA via self-training has gained a lot of attention~\cite{yang2021st3d, saltori2020SFUDA3DSU, you2021exploiting}.
Similar to 2D approaches~\cite{RoyChowdhury2019autoadaptod, cai2019meanteacher, khodabandeh2019robustdaod}, the idea is to use a 3D detector pre-trained on a labelled source dataset and apply it on the target dataset to obtain pseudo-labels.
These labels are leveraged to re-train the model.
Both steps, label generation and re-training, are repeated until convergence.
However, generating reliable pseudo-labels is a non-trivial task.

Although most real-world data is continuous in nature, this property is rarely exploited for UDA of 3D object detectors. 
As a notable exception,~\cite{you2021exploiting} leverages a probabilistic offline tracker.
Though simple and effective, a major weakness of probabilistic trackers is that they heavily depend on the detection quality.
Because of the domain gap, however, the detection quality usually degrades significantly.
Additionally, these trackers require a hand-crafted motion model which must be adjusted manually.
These limitations result in unreliable pseudo-labels driven by missing and falsely classified objects.

\begin{figure}[t!]
	\centering
	\includegraphics[width=1.0\textwidth]{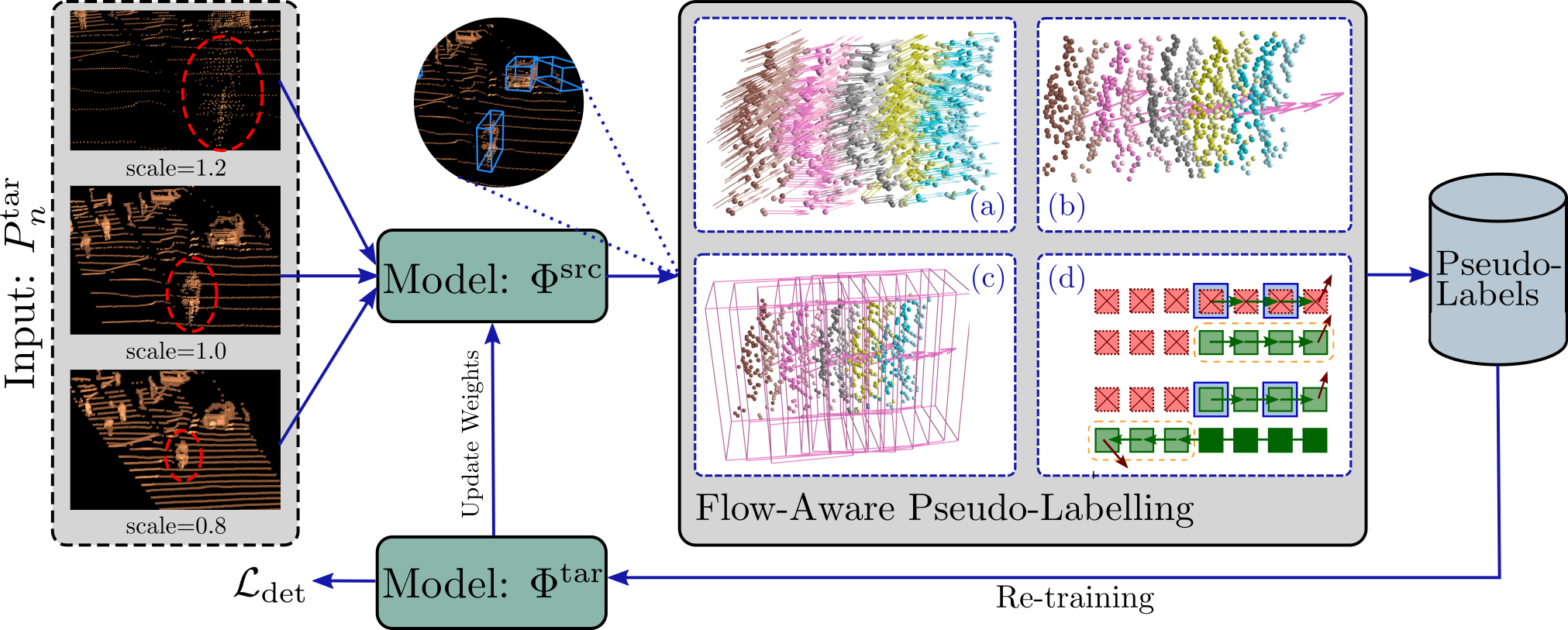}
	\caption{FAST3D scheme. The initial model $ \Phi^{\text{src}} $, pre-trained on source data, generates non-confident detections on the target data for different scales.
		We leverage scene flow information (a) to calculate box flow (b), \ie the mean flow of an object. This allows us to increase the recall and propagate labels through time (c).
		A final refinement step (d) reduces false positives and false negatives to obtain high-quality pseudo-labels for re-training.
		Points with the same colour originate from the same point cloud $ P^{\text{tar}}_n $.}
	\label{fig:intro_overview}
\end{figure}

To overcome these issues, we present our \textbf{F}low-\textbf{A}ware \textbf{S}elf-\textbf{T}raining approach for \textbf{3D} object detection (\emph{FAST3D}), leveraging scene flow~\cite{liu2019flownet3d, liu2019meteornet, wu2020pointpwc} for robust pseudo-labels, as illustrated in Fig.~\ref{fig:intro_overview}.
Scene flow estimates per-point correspondences between consecutive point clouds and calculates the motion vector for each point.
Furthermore, although trained on synthetic data~\cite{mayer2016flyingthings} only, scene flow estimators already achieve a favourable accuracy on real-world road data~\cite{wu2020pointpwc}.
Thus, we investigate scene flow for UDA.
In particular, we will show that scene flow allows us to propagate pseudo-labels reliably, recover missed objects and discard unreliable detections.
This results in significantly improved pseudo-label quality which boosts the performance of self-training 3D detectors drastically.

We conduct experiments on the challenging Waymo Open Dataset (WOD)~\cite{sun2020waymo} considering two state-of-the-art 3D detectors, PointRCNN~\cite{shi2018pointrcnn} and PV-RCNN~\cite{shi2020pvrcnn}, pre-trained on the much smaller KITTI dataset~\cite{geiger2012ad, geiger2013vision}.
Without any prior target domain knowledge (\eg~\cite{wang2020train}), nor the need for source domain data (\eg~\cite{yang2021st3d}), we surpass the state-of-the-art by a significant margin.

\section{Related Work}
\paragraph{3D Object Detection} 
3D object detectors localize and classify an unknown number of objects within a 3D environment. Commonly, the identified objects are represented as tightly fitting oriented bounding boxes. Most recent 3D detectors, trained and evaluated on autonomous driving datasets, operate on LiDAR point clouds. 

One way to categorize them is by their input representation. Voxel-based approaches~\cite{lang2019pointpillars, yan2018second, li20173dfully, wang2020pillarbased, bin2018pixor, zhou2018voxelnet, Ye2020HVnet, deng2020voxelrcnn} rasterize the input space and assign points of irregular and sparse nature to grid cells of a fixed size. Afterwards, they either project these voxels directly to the bird's eye view (BEV) or first learn feature representations by leveraging 3D convolutions and project them to BEV afterwards. Finally, a conventional 2D detection head predicts bounding boxes and class labels.
Another line of work are point-based detectors~\cite{ngiam2019starnet, shi2020pointgnn, yang20203dssd, shi2018pointrcnn}. In order to generate proposals directly from points, they leverage PointNet~\cite{qi2017pointnet, qi2017pointnetpp} to extract point-wise features. Hybrid approaches~\cite{chen2019fastpointrcnn, shi2020points, shi2020pvrcnn, he2020sassd, yang2019std, zhou2020mvf}, on the other hand, seek to leverage the advantages of both of the aforementioned strategies. In contrast to point cloud-only approaches, multi-modal detectors~\cite{chen2017mv3d, ku2018avod, liang2019mtms, liang2018deepcontfuse, xu2018pointfusion, qi2017frustum} utilize 2D images complementary to LiDAR point clouds. The additional image information can be beneficial to recognise small objects. 

Since most state-of-the-art approaches operate on LiDAR point clouds only, we also focus on this type of detectors.
In particular, we demonstrate our self-training approach using PointRCNN~\cite{shi2018pointrcnn} and PV-RCNN~\cite{shi2020pvrcnn}.
Both detectors have already been used in UDA settings for self-driving vehicles~\cite{wang2020train, yang2021st3d} and achieve state-of-the-art robustness and accuracy. 

\paragraph{Scene Flow}
Scene flow represents the 3D motion field in the scene~\cite{vedula1999sceneflow}. Recently, a relatively new area of research aims to estimate point-wise motion predictions in an end-to-end manner directly from raw point clouds~\cite{liu2019flownet3d, gu2019hplflownet, wu2020pointpwc}. With few exceptions~\cite{liu2019meteornet}, most approaches process two consecutive point clouds as input. A huge benefit of data-driven scene flow models, especially regarding UDA, is their ability to learn in an unsupervised manner~\cite{mittal2020justgowith, wu2020pointpwc, li2021selfpointflow}. The biggest drawback of these networks is their huge memory consumption which limits the number of input points. To address this,~\cite{jund20201scaleablesceneflow} proposes a light-weight model applicable to the dense Waymo Open Dataset~\cite{sun2020waymo} point clouds.

In our work, we use the 3D motion field to obtain robust and reliable pseudo-labels for UDA by leveraging the motion consistency of sequential detections.
To this end, we utilize PointPWC~\cite{wu2020pointpwc} which achieves state-of-the-art scene flow estimation performance.

\paragraph{Unsupervised Domain Adaptation (UDA)}
The common pipeline for UDA is to start with a model trained in a fully supervised manner on a label-rich source domain and adapt it on data from a label-scarce target domain in an unsupervised manner.
Hence, the goal is to close the gap between both domains.
There is already a large body of literature on UDA for 2D object detection in driving scenes~\cite{chen2018domain, he2019MultiAdversarialFF, hsu2020progressivedet, kim2019diversify, rodriguez2019domainAF, saito2019strongweak, zhuang2020iFANIF, wang2019universaldet, xu2020ExploringCR}.
Due to the growing number of publicly available large-scale autonomous driving datasets, UDA on 3D point cloud data has gained more interest recently~\cite{qin2019PointDANAM, jaritz2019xmuda, Yi2020CompleteL, achituve2021self}.

For LiDAR-based 3D object detection, \cite{wang2020train} demonstrate serious domain gaps between various datasets, mostly caused by different sensor setups (\eg resolution or mounting position) or geographic locations (\eg Germany $\to$ USA).
A popular solution to UDA is self-training, as for the 2D case~\cite{cai2019meanteacher, khodabandeh2019robustdaod, RoyChowdhury2019autoadaptod}.
For example,~\cite{yang2021st3d} initially train the detector with random object scaling and an additional score prediction branch.
Afterwards, pseudo-labels are updated in a cyclic manner considering previous examples.
In order to overcome the need for source data,~\cite{saltori2020SFUDA3DSU} performs test-time augmentation with multiple scales. The best matching labels are selected by checking motion coherency. 

In contrast to~\cite{you2021exploiting}, where temporal information is exploited by probabilistic tracking, we propose to leverage scene flow.
This enables us to reliably extract pseudo-labels despite initially low detection quality by exploiting motion consistency.
As in~\cite{saltori2020SFUDA3DSU}, we also utilize test-time augmentation to overcome scaling issues, but we only need two additional scales.

\section{Flow-Aware Self-Training}
We now introduce our Flow-Aware Self-Training approach \emph{FAST3D}, which consists of four steps as illustrated in Fig.~\ref{fig:intro_overview}.
First, starting with a model trained on the source domain, we obtain initial 3D object detections for sequences of the target domain (Sec.~\ref{sec:approach-generation}).
Second, we leverage scene flow to propagate these detections throughout the sequences and obtain tracks, robust to the initial detection quality (Sec.~\ref{sec:approach-tracker}).
Third, we recover potentially missed tracks and correct false positives in a refinement step (Sec.~\ref{sec:label_refinement}).
Finally, we extract pseudo-labels for self-training to improve the initial model (Sec.~\ref{sec:approach-retraining}).

\paragraph{Problem Statement}
Given a 3D object detection model $ \Phi^{\text{src}} $ pre-trained on $ N^{\text{src}} $ source domain frames $ \{(P^{\text{src}}_n, L^{\text{src}}_n)\}_{n=1}^{N^{\text{src}}} $, our task is to adapt to unseen target data with $ N^{\text{tar}} $ unlabelled sequences $ \{S^{\text{tar}}_i\}_{i=1}^{N^{\text{tar}}} $, where $ S^{\text{tar}}_i = \{P^{\text{tar}}_n\}_{n=1}^{N_i^{\text{tar}}} $ with varying length $N_i^{\text{tar}}$.
Here, $ P_n $ and $ L_n $ denote the point cloud and corresponding labels of the $ n^{\text{th}} $ frame.
To obtain the target detection model $\Phi^{\text{tar}}$, we apply self-training which assumes that both domains contain the same set of classes (\ie vehicle/car, pedestrian, cyclist) but these are drawn from different distributions.
In the following, we show how to self-train a detection model $\Phi^{\text{tar}}$ without access to source data or target label statistics (as \eg~\cite{wang2020train}) and without modifying the source model in any way (as \eg~\cite{yang2021st3d}) to achieve performance beyond the state-of-the-art.
Because we only work with target domain data, we omit the domain superscripts to improve readability.

\subsection{Pseudo-Label Generation}
\label{sec:approach-generation}
The first step, as in all self-training approaches (\eg~\cite{yang2021st3d, you2021exploiting, saltori2020SFUDA3DSU}), is to run the initial source model $\Phi^{\text{src}}$ on the target data and keep high-confident detections.
More formally, the $j^\text{th}$ detection at frame $n$ is represented by its bounding box $\vec{b}_{j,n} = \left(\vec{p}, \vec{d}, \theta\right)$ and confidence score $c_{j,n}^\vec{b}$.
Each box is defined by its centre position $ \vec{p}=(c_x, c_y, c_z) $, dimension $ \vec{d} = ( l, w, h) $ and heading angle $ \theta $.
Note that we can use any vanilla 3D object detector, since we do not change the initial model at all, not even by adapting the mean object size of anchors (as in~\cite{yang2021st3d}).
In order to deal with different object sizes between various datasets without pre-training $\Phi^{\text{src}}$ on scaled source data~\cite{yang2021st3d, wang2020train}, we leverage test-time augmentation instead.
In particular, we feed $\Phi^{\text{src}}$ with three scales (\ie~0.8, 1.0, 1.2) of the same point cloud, similar to~\cite{saltori2020SFUDA3DSU}.
However,~\cite{saltori2020SFUDA3DSU} requires to run the detector on \textbf{125} different scale levels (due to different scaling factors along each axis).
By leveraging scene flow, we only need three scales to robustly handle both larger and smaller objects.
We combine the detection outputs at all three scales via non-maximum suppression (NMS) and threshold the detections at a confidence of $ c_{j,n}^\vec{b} \geq 0.8 $ to obtain the initial high-confident detections.

\subsection{Flow-Aware Pseudo-Label Propagation}
\label{sec:approach-tracker}

Keeping only high-confident detections of the source model gets rid of false positives (FP) but also results in a lot of false negatives (FN).
While this can partially be addressed via multi-target tracking (MTT), \eg~\cite{you2021exploiting}, standard MTT approaches (such as~\cite{weng2020AB3DMOT}) are not suitable for low quality detections because of their hand-crafted motion prediction.
Due to the domain gap, however, the detection quality of the source model will inevitably be low.

To overcome this issue, we introduce our scene flow-aware tracker.
We define a set of tracks $ T=\{t_1, t_2,\dots, t_k\} $ where each track $ t_k $ contains a list of bounding boxes and is represented by its current state $ \vec{x}_{k,n} $, \ie its bounding box, and track confidence score $c_{k,n}^\vec{t}$.
Following the tracking-by-detection paradigm, we use the high-confident detections and match them in subsequent frames.
Instead of hand-crafted motion models, however, we utilize scene flow to propagate detections. 
More formally, given two consecutive point clouds $ P_{n-1} $ and $ P_{n} $, PointPWC~\cite{wu2020pointpwc} estimates the motion vector $ \vec{v}_{i,n-1} = (v_x, v_y, v_z)_{i,n-1} $ for each point $ \vec{p}_{i,n-1} \in P_{n-1} $. 
We then average all motion vectors within a track's bounding box $ \vec{x}_{k,n-1} $ to compute its mean flow $ \vec{v}_{k,n-1} $. 
To get the predicted state $ \vec{x}_{k,n|n-1} $ for the current frame $n$, we estimate the position of each track as $ \vec{p}_{k,n|n-1} = \vec{p}_{k,n-1} + \vec{v}_{k,n-1} $. 
We then assign detections to tracks via the Hungarian algorithm~\cite{Kuhn55thehungarian}, where we use the intersection over union between the detections $\vec{b}_{j,n}$ and the predicted tracks $\vec{x}_{k,n|n-1}$.

We initialize a new track for each unassigned detection.
This naive initialisation ensures that we include all object tracks, whereas potentially wrong tracks can easily be filtered in our refinement step (Sec.~\ref{sec:label_refinement}). Initially, we set a track's confidence $c_{k,n}^\vec{t}$ to the detection confidence $c_{j,n}^\vec{b}$.
For each assigned pair, we perform a weighted update based on the confidence scores as $ \vec{x}_{k,n|n} = \frac{c_{k,n-1}^\vec{t} \vec{x}_{k,n|n-1} + c_{j,n}^\vec{b} \vec{b}_{j,n}}{c_{k,n-1}^\vec{t} + c_{j,n}^\vec{b}} $ and update the track confidence $ c_{k,n}^\vec{t} = \frac{(c_{k,n-1}^\vec{t})^2 + (c_{j,n}^\vec{b})^2}{c_{k,n-1}^\vec{t} + c_{j,n}^\vec{b}} $.
Note that we update the track's heading angle only if the orientation change is less than $30^\circ$, otherwise we keep its previous orientation.
For boxes with only very few point correspondences, the flow estimates may be noisy.
To suppress such flow outliers, we allow a maximum change in velocity of $ 1.5~\text{m/s}$, along with the maximum orientation change of $30^\circ$.
If the estimated flow exceeds these limits, we keep the previous estimate.

For track termination, we distinguish moving (\ie $\vec{v}_{k,n-1} > 0.8~\text{m/s}$) and static objects.
For moving objects, we terminate their tracks if there are no more points within their bounding box.
Static object tracks are kept alive as long as they are within the field-of-view.
This ensures that even long term occlusions can be handled, as such objects will eventually contain points or be removed during refinement.

\begin{figure}[t!]
	\centering
	\subfigure[Recovery via flow consistency.]{\label{fig:track_recovery_consistency}{\includegraphics[width=0.49\textwidth]{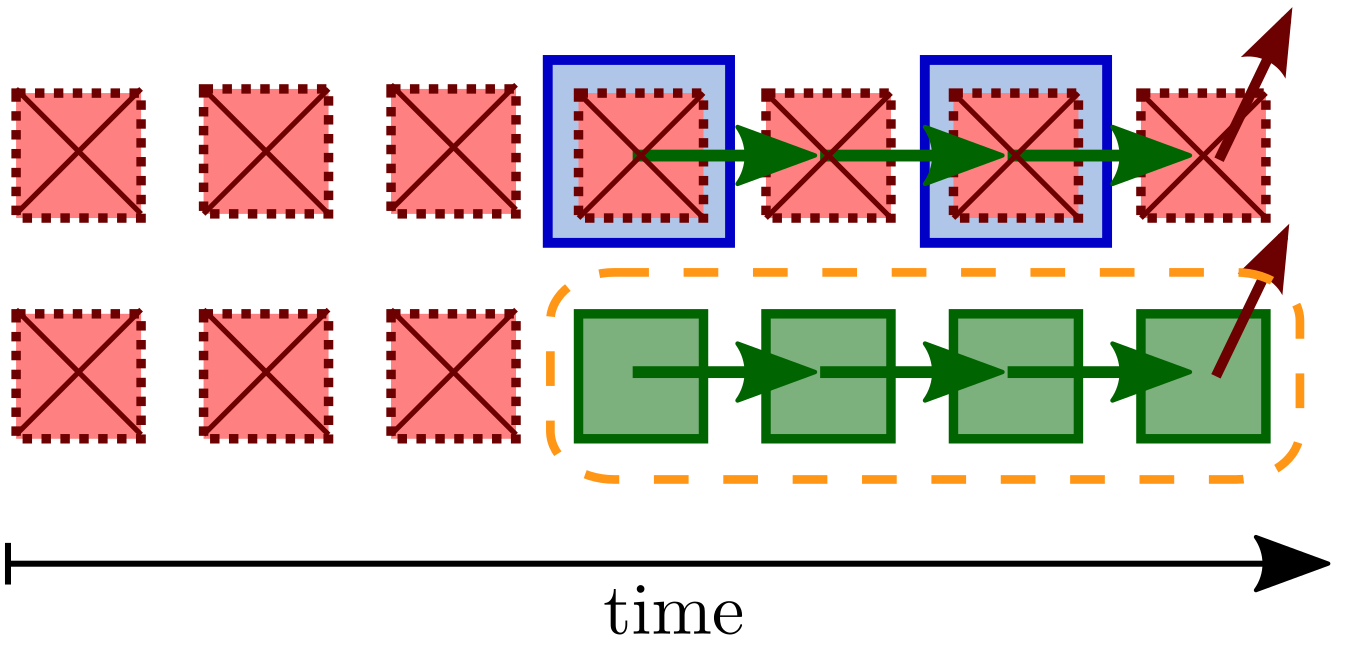}}}
	\hfill
	\subfigure[Backward tracking.]{\label{fig:track_recovery_tail}{\includegraphics[width=0.49\textwidth]{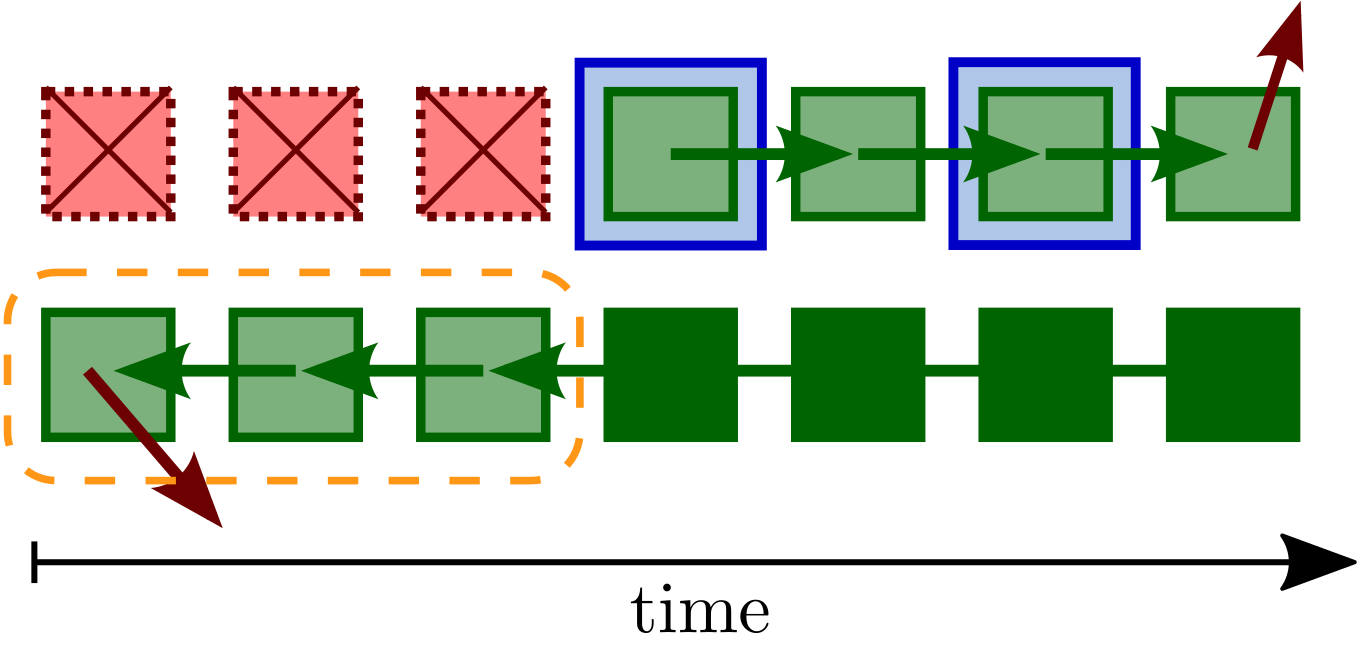}}}
	\caption{Track refinement to address the initially low detection quality on the target domain. The valid track, false negative (missed) and true positive detections are illustrated in green, red and blue, respectively. Green and red arrows show valid and inconsistent flow (high change in orientation/velocity), respectively. (a) Flow consistency recovers object tracks with low detection hit ratio. (b) Starting from the first detection, backward tracking propagates the bounding box back in time to recover missing tails. Refined tracks are marked by the dashed orange regions.}
	\label{fig:track_recovery}
\end{figure}

\subsection{Pseudo-Label Refinement}
\label{sec:label_refinement}
Due to our simplified track management (initialisation and termination), we now have tracks covering most of the visible objects but suffer from several false positive tracks.
However, these can easily be corrected in the following refinement step.

\paragraph{Track Filtering and Correction}
We first compute the \emph{hit ratio} for each track, \ie the number of assigned detections divided by the track's length.
Tracks are removed if their hit ratio is less than $0.3$.
Additionally, we discard tracks shorter than $0.5~\text{s}$ (\ie 5 frames at the typical LiDAR frequency of $10~\text{Hz}$) and tracks which detections never exceed 15 points (to suppress spurious detections).

In our experiments, we observed that detections are most accurate on objects with more points (as opposed to the confidence score which is usually less reliable due to the domain gap).
Thus, we sort a track's assigned detections by the number of contained points and compute the average dimension over the top three.
We then apply this average dimension to all boxes of the track.
Additionally, for static cars we also apply the average position and heading angle to all boxes of the track since these properties cannot change for rigid static objects.

\paragraph{Track Recovery} This conservative filtering of unreliable tracks ensures a lower number of false positives. However, true positive tracks with only very few detections might be removed prematurely. In order to recover these, we leverage \emph{flow consistency}.
As illustrated in Fig.~\ref{fig:track_recovery_consistency}, we consider removed tracks with at least two non-overlapping detections (\ie moving objects).
% This ensures that we deal with a target of recognizable motion.
For these tracks, we propagate the first detection by solely utilizing the scene flow estimates.
If this flow-extrapolated box overlaps with the other detections (minimum IoU of $0.3$), we recover this track.

\paragraph{Backward Completion} A common drawback of the tracking-by-detection paradigm is late initialization caused by missing detections at the beginning of the track. To overcome this issue, we look backwards for each track as illustrated in Fig.~\ref{fig:track_recovery_tail} to extend missing tails. Hence, we leverage scene flow in the opposite direction and propagate the bounding box back in time. We propagate backwards as long as the flow is consistent, meaning no unreasonable changes in velocity or direction, and the predicted box location contains points.
To avoid including the ground plane in the latter condition, we count only points within the upper $70\,\%$ of the bounding box volume.

\subsection{Self-Training}
\label{sec:approach-retraining}
We use the individual bounding boxes of the refined tracks as high-quality pseudo-labels to re-train the detection model.
To this end, we use standard augmentation strategies, \ie~randomized world scaling, flipping, rotation, as well as ground truth sampling~\cite{shi2020pvrcnn,yan2018second}.
As we don't need to modify the initial model $ \Phi^{\text{src}} $, we also use its detection loss $ \mathcal{L}_{\text{det}} $ without modifications for re-training.

\section{Experiments}
In the following, we present the results of our flow-aware self-training approach using two vanilla 3D object detectors.
Additional evaluations demonstrating the improved pseudo-label quality are included in the supplemental material.

\subsection{Datasets and Evaluation Details}

\paragraph{Datasets}
We conduct experiments on the challenging Waymo Open Dataset (WOD)~\cite{sun2020waymo} with a source model $ \Phi^s $ pre-trained on the KITTI dataset~\cite{geiger2012ad, geiger2013vision}.
We sample 200 sequences ($ \sim $25\%) from the WOD training set for pseudo-label generation and 20 sequences ($ \sim $10\%) from the WOD validation set for testing.
With this evaluation, we cover various sources of domain shifts: geographic location (Germany~$ \rightarrow $~USA), sensor setup (Velodyne~$ \rightarrow $~Waymo LiDAR) and weather (broad daylight~$ \rightarrow $~sunny, rain, night). In contrast to recent work, we do not only consider the car/vehicle class, but also pedestrians and cyclists.
Because the initial model $ \Phi^s $ is trained on front view scans only (available KITTI annotations), we stick to this setup for evaluation.

\paragraph{Evaluation Metrics}
We follow the official WOD evaluation protocol~\cite{sun2020waymo} and report the average precision (AP) for the intersection over union (IoU) thresholds $ 0.7 $ and $ 0.5 $ for vehicles, as well as $ 0.5 $ and $ 0.25 $ for pedestrians and cyclists.
AP is reported for both bird's eye view (BEV) and 3D -- denoted as $ \text{AP}_{\text{BEV}} $ and $ \text{AP}_{\text{3D}} $ -- for different ranges: $ 0 \,\text{m}-30 \,\text{m} $, $ 30 \,\text{m}-50 \,\text{m} $ and $ 50 \,\text{m}-75 \,\text{m} $. Additionally, we calculate the \emph{closed gap} as in~\cite{yang2021st3d}. 

\paragraph{Comparisons}
The lower performance bound is to directly apply the source model $ \Phi^s $ on the target data, denoted source only (SO).
The fully supervised model (FS) trained on the whole target data ($ \sim 25\% $~of WOD) on the other hand defines the upper performance bound. 
Statistical normalization (SN)~\cite{wang2020train} is the state-of-the-art on UDA for 3D object detection and leverages target data statistics.
The recent DREAMING~\cite{you2021exploiting} approach also exploits temporal information for UDA.
Finally, we also compare against the few shot~\cite{wang2020train} approach which adapts the model using a few fully labelled sequences of the target domain.

\paragraph{Implementation details}
We demonstrate FAST3D with two detectors, PointRCNN~\cite{shi2018pointrcnn} and PV-RCNN~\cite{shi2020pvrcnn} from OpenPCDet~\cite{openpcdet2020}, configured as in their official implementations.
To compensate for the ego-vehicle motion, we project all frames of a sequence into the world coordinate system by computing the ego-vehicle poses~\cite{sun2020waymo}.
We slightly increase the field-of-view (FOV) from approximately $ 45^{\circ} $ to $ 60^{\circ} $, which simplifies handling pseudo-label tracks at the edge of the narrower FOV.
In each self-training cycle, we re-train the detector for $ 4 $ epochs with Adam~\cite{diderik2015adam} and a learning rate of $ 3\times 10^{-4} $ until we reach convergence (\ie PointRCNN 2 cycles, PV-RCNN 3 cycles).

We estimate the scene flow using PointPWC~\cite{wu2020pointpwc}. 
Although PointPWC, similar to other flow estimators~\cite{li2021selfpointflow, mittal2020justgowith}, could be fine-tuned in a self-supervised fashion, we use the off-the-shelf model pre-trained only on synthetic data~\cite{mayer2016flyingthings}.
This allows us to demonstrate the benefits and simplicity of leveraging 3D motion without additional fine-tuning.
To prepare the point clouds for scene flow estimation, we use RANSAC~\cite{fischler1981ransac} to remove ground points (as in~\cite{wu2020pointpwc}) and randomly subsample the larger input point cloud to match the smaller one.

\subsection{Empirical Results}
In Table~\ref{tab:fast3d_overall} we compare FAST3D for two detectors with the source only~(SO) and fully supervised~(FS) strategies, which define the lower and upper performance bound, respectively.
Across all classes and different IoU thresholds, we manage to close the domain gap significantly (36\% to 87\%), achieving almost the fully supervised oracle performance although our approach is unsupervised and does not rely on any prior knowledge about the target domain.
To the best of our knowledge, we are the first to additionally report adaptation performance for both pedestrians and cyclists as these vulnerable road users should not be neglected in evaluations. 
We focus this evaluation on $\text{AP}_{\text{3D}}$ because (in contrast to $\text{AP}_{\text{BEV}}$) this metric also penalizes estimation errors along the $z$-axis (\ie vertical centre and height).
Consequently, $\text{AP}_{\text{BEV}}$ scores are usually much higher than $\text{AP}_{\text{3D}}$.

Since reporting the closed domain gap has only been proposed very recently~\cite{yang2021st3d}, no other approaches reported this closure for KITTI$\rightarrow$WOD yet.
We can, however, relate the results for the most similar setting, \ie adaptation of PV-RCNN for the car/vehicle class, where we achieve a closed gap of 76.5\% (KITTI$\rightarrow$WOD, \emph{vehicles} within all sensing ranges) in contrast to 70.7\% of~\cite{yang2021st3d} (WOD$\rightarrow$KITTI, \emph{cars} within KITTI's \emph{moderate} difficulty).
Despite the different settings, we achieve a favourable domain gap closure while evaluating on the more challenging KITTI$\rightarrow$WOD.
In particular, according to~\cite{wang2020train}, starting from KITTI as the source domain is the most difficult setting for UDA, as it has orders of magnitude fewer samples than other datasets.
Additionally, KITTI contains only sunny daylight data and its \emph{car} class does not include trucks, vans or buses which are, however, contained in the WOD \emph{vehicle} class.

\begin{table}[ht]
	\setlength\tabcolsep{4pt}
	\footnotesize
	\bgroup
	\def\arraystretch{1.2}
	\begin{center}
		\begin{tabular}{ c c l | c c c | c || c c c | c }
			\cline{3-11}
			\multicolumn{2}{c}{} & \multirow{2}{*}{Class} & \multicolumn{4}{|c||}{PointRCNN} & \multicolumn{4}{c}{PV-RCNN} \\
			\cline{4-11}
			\multicolumn{3}{c|}{} & \textbf{SO}  & \textbf{FAST3D} & \textbf{FS} & \textbf{Closed Gap} & \textbf{SO}  & \textbf{FAST3D} & \textbf{FS}  & \textbf{Closed Gap} \\
			\cline{1-11}
			\multirow{3}{*}{\rotatebox[origin=c]{90}{\textbf{AP\textsubscript{3D}}}} 
			& \rotatebox[origin=c]{90}{\tiny{0.7}} &
			Vehicle    & \phantom{0}8.0 & 58.1 & 65.6 & 87.0 \% & 10.3 & 63.6 & 80.0 & 76.5 \% \\
			\cline{3-11}
			& \rotatebox[origin=c]{90}{\tiny{0.5}} & Pedestrian & 18.9 & 36.8 & 47.8 & 62.0 \% & 13.7 & 44.7 & 74.4 & 51.1 \% \\
			\cline{3-11}
			& \rotatebox[origin=c]{90}{\tiny{0.5}} & Cyclist    & 24.2 & 48.0 & 60.0 & 66.5 \% & 14.6 & 35.6 & 72.9 &  36.0 \%\\
			\cline{1-11}
			\noalign{\vskip\arrayrulewidth \vskip\doublerulesep}
			\cline{1-11}
			\multirow{3}{*}{\rotatebox[origin=c]{90}{\textbf{AP\textsubscript{3D}}}} 
			& \rotatebox[origin=c]{90}{\tiny{0.5}} &
			Vehicle    & 60.0 & 77.3 & 80.9 & 82.8 \% & 50.3 & 85.1 & 95.2 & 77.5 \%\\
			\cline{3-11}
			& \rotatebox[origin=c]{90}{\tiny{0.25}} & Pedestrian & 30.8 & 41.7 & 52.2 & 50.9 \% & 21.6 & 57.2 & 85.4 & 55.8 \%\\
			\cline{3-11}
			& \rotatebox[origin=c]{90}{\tiny{0.25}} & Cyclist    & 31.8 & 66.2 & 69.0 & 92.5 \% & 32.5 & 49.4 & 75.5 & 39.3 \%\\
			\cline{1-11}
		\end{tabular}
	\end{center}
	\egroup
	\caption{Our FAST3D compared to lower (source-only, SO) and upper (fully-supervised, FS) bound, on KITTI $\rightarrow$ Waymo Open Dataset (WOD) for different IoU thresholds.}
	\label{tab:fast3d_overall}
\end{table}

Table~\ref{tab:fast3d_ranges} reports the results in $\text{AP}_{\text{3D}} $ split by their detection range for different intersection over union (IoU) thresholds.
We can observe that our self-training pipeline significantly improves the initial model on all ranges.
The only notable outlier are cyclists within the far range of 50--75 meters, where even after re-training both detectors achieve rather low scores.
This is due to the low number of far range cyclists within the WOD validation set, \ie only a few detection failures already drastically degrade the $\text{AP}_{\text{3D}} $ for the \emph{cyclist} class.

\begin{table}[ht]
	\setlength\tabcolsep{4pt}
	\footnotesize
	\bgroup
	\def\arraystretch{1.2}
	\begin{center}
		\begin{tabular}{ c c l | c | c | c || c | c | c }
			\cline{3-9}
			\multicolumn{2}{c}{} & \multirow{2}{*}{Class} & \multicolumn{3}{|c||}{PointRCNN} & \multicolumn{3}{c}{PV-RCNN} \\
			\cline{4-9}
			\multicolumn{3}{c|}{}  & \textbf{0m~-~30m} & \textbf{30m~-~50m} & \textbf{50m~-~75m} & \textbf{0m~-~30m} & \textbf{30m~-~50m} & \textbf{50m~-~75m} \\
			\cline{1-9}
			\multirow{3}{*}{\rotatebox[origin=c]{90}{\textbf{AP\textsubscript{3D}}}} 
			& \rotatebox[origin=c]{90}{\tiny{0.7}} &
			Vehicle     & 70.3 & 58.7 & 37.0 & 74.9 & 64.6 & 44.0\\
			\cline{3-9}
			& \rotatebox[origin=c]{90}{\tiny{0.5}} & 
			Pedestrian  & 66.2 & 45.7 & 14.6 & 66.5 & 55.0 & 22.7 \\
			\cline{3-9}
			& \rotatebox[origin=c]{90}{\tiny{0.5}} & 
			Cyclist     & 76.7 & 65.7 & \phantom{0}4.9 & 64.4 & 46.0 & \phantom{0}1.4\\
			\cline{1-9}
			\noalign{\vskip\arrayrulewidth \vskip\doublerulesep}
			\cline{1-9}
			\multirow{3}{*}{\rotatebox[origin=c]{90}{\textbf{AP\textsubscript{3D}}}} 
			& \rotatebox[origin=c]{90}{\tiny{0.5}} &
			Vehicle     & 86.9 & 79.3 & 63.5 & 91.2 & 85.9 & 70.4\\
			\cline{3-9}
			& \rotatebox[origin=c]{90}{\tiny{0.25}} & 
			Pedestrian  & 74.1 & 50.3 & 17.4 & 78.0 & 67.3 & 36.4\\
			\cline{3-9}
			& \rotatebox[origin=c]{90}{\tiny{0.25}} & 
			Cyclist     & 93.9 & 76.2 & 33.6 & 90.6 & 54.4 & \phantom{0}5.0\\
			\cline{1-9}
		\end{tabular}
	\end{center}
	\egroup
	\caption{Detailed results at different sensing ranges (KITTI $\rightarrow$ WOD) for two detectors at different IoU thresholds.}
	\label{tab:fast3d_ranges}
\end{table}

\subsection{Comparison with the State-of-the-Art}
To fairly compare with the state-of-the-art in Table~\ref{tab:fast3d_compare}, we follow the common protocol and report both $\text{AP}_{\text{3D}} $ and $\text{AP}_{\text{BEV}} $ on the \emph{vehicle} class.
We clearly outperform the best approach (statistical normalization), even though this method utilizes target data statistics and is thus considered weakly supervised, whereas our approach is unsupervised. We also outperform the only temporal 3D pseudo-label approach~\cite{you2021exploiting} by a huge margin for all ranges, especially at the high-quality $\text{IoU}\geq 0.7$. Moreover, we perform on par with the few shot approach from~\cite{wang2020train} on close-range data and even outperform it on medium and far ranges. 

\begin{table}[ht]
	\setlength\tabcolsep{4pt}
	\footnotesize
	\bgroup
	\def\arraystretch{1.2}
	\begin{center}
		\begin{tabular}{ l | c | c | c || c | c | c }
			\cline{1-7}
			\multirow{2}{*}{Method} & \multicolumn{3}{|c||}{IoU 0.7} & \multicolumn{3}{c}{IoU 0.5} \\
			\cline{2-7}
			& \textbf{0m~-~30m} & \textbf{30m~-~50m} & \textbf{50m~-~75m} & \textbf{0m~-~30m} & \textbf{30m~-~50m} & \textbf{50m~-~75m} \\
			\cline{1-7}
			SO     & 29.2 / 10.0 & 27.2 / ~8.0 & 24.7 / ~4.2 & 67.8 / 66.8 & 70.2 / 63.9 & 48.0 / 38.5 \\
			\cline{1-7}
			SN~\cite{wang2020train} & \textbf{87.1} / 60.1 & \textbf{78.1} / 54.9 & 46.8 / 25.1  & - & - & - \\
			\cline{1-7}
			DREAMING~\cite{you2021exploiting} & 51.4 / 13.8 & 44.5 / 16.7 & 25.6 / ~7.8 & 81.1 / \textbf{78.5} & 69.9 / 61.8 & 50.2 / 41.0\\
			\cline{1-7}
			FAST3D (ours) & 81.7 / \textbf{70.3} & 75.4 / \textbf{58.7} & \textbf{52.4} / \textbf{37.0} & \textbf{87.2} / 74.9 & \textbf{81.2} / \textbf{64.6} & \textbf{68.6} / \textbf{44.0} \\
			\cline{1-7}
			\noalign{\vskip\arrayrulewidth \vskip\doublerulesep}
			\cline{1-7}
			Few Shot~\cite{wang2020train} & 88.7 / 74.1 & 78.1 / 57.2 & 45.2 / 24.3  & - & - & - \\
			\cline{1-7}
			\noalign{\vskip\arrayrulewidth \vskip\doublerulesep}
			\cline{1-7}
			FS    & 86.2 / 78.5 & 77.8 / 63.7 & 60.8 / 48.1 & 91.7 / 92.1 & 83.7 / 78.3 & 73.3 / 69.7\\
			\cline{1-7}
		\end{tabular}	
	\end{center}
	\egroup
	\caption{Comparison to the state-of-the-art for KITTI $\rightarrow$ WOD. All approaches adapt PointRCNN. Following the common protocol, results are listed for the \emph{vehicle} class at different sensing ranges reported in $\text{AP}_{\text{BEV}} $ / $\text{AP}_{\text{3D}} $.}
	\label{tab:fast3d_compare}
\end{table}

\section{Conclusion}
We presented a flow-aware self-training pipeline for unsupervised domain adaptation of 3D object detectors on sequential LiDAR point clouds.
Leveraging motion consistency via scene flow, we obtain reliable and precise pseudo-labels at high recall levels.
We do not exploit any prior target domain knowledge, nor do we need to modify the 3D detector in any way.
As demonstrated in our evaluations, we surpass the current state-of-the-art in self-training for UDA by a large margin.

\section*{Acknowledgements}
The financial support by the Austrian Federal Ministry for Digital and Economic Affairs, the National Foundation for Research, Technology and Development and the Christian Doppler Research Association is gratefully acknowledged. 

\bibliography{abbreviations_short,egbib}
\end{document}